# Parallel Intent and Slot Prediction using MLB Fusion

**Anmol Bhasin\*, Bharatram Natarajan\*, Gaurav Mathur\* and Himanshu Mangla\***

Samsung R&D Institute, India - Bangalore

anmol.bhasin@samsung.com, bharatram.n@samsung.com, gaurav.m4@samsung.com, him.mangla@gmail.com

*Abstract*—**Intent and Slot Identification are two important tasks in Spoken Language Understanding (SLU). For a natural language utterance, there is a high correlation between these two tasks. A lot of work has been done on each of these using Recurrent-Neural-Networks (RNN), Convolution Neural Networks (CNN) and Attention based models. Most of the past work used two separate models for intent and slot prediction. Some of them also used sequence-to-sequence type models where slots are predicted after evaluating the utterance-level intent. In this work, we propose a parallel Intent and Slot Prediction technique where separate Bidirectional Gated Recurrent Units (GRU) are used for each task. We posit the usage of MLB (Multimodal Low-rank Bilinear Attention Network) fusion for improvement in performance of intent and slot learning. To the best of our knowledge, this is the first attempt of using such a technique on text based problems. Also, our proposed methods outperform the existing state-of-the-art results for both intent and slot prediction on two benchmark datasets**

## I. INTRODUCTION

Conversational Agents (CA) like Samsung's Bixby, Amazon's Alexa, and Microsoft's Cortana are becoming increasingly popular [1]. The primary input for these Conversational Agents (CA) is human voice. This human speech is converted into text using Automatic Speech Recognition systems. This text, then goes to a Natural Language Understanding (NLU) software module. This NLU module has three primary sub-tasks namely Domain Classification, Intent Classification and Slot Labelling. The process of classifying the NL input into top-level classes (e.g. Phone, Calendar, Music etc.) is called as Domain Classification. Identifying the action (e.g. "placing a call", "setting a reminder" or "playing a song" etc.) that the user wants to perform is called Intent Classification. Extracting the attributes (e.g. name of a person, time, song name etc.) from the input falls under the purview of Slot Labelling (SL). A lot of work has been pursued previously in each of these areas [2-8].

Of these three sub tasks, our work focusses on joint intent and slot prediction technique by establishing a relationship between the two through MLB fusion and Dense addition. Let's take a sample utterance from ATIS dataset (Table 1) "Show flights from Washington to San Francisco between 6pm and 8pm on Friday". In this utterance, user's intention is to fly (intent) and he wants to fly from Washington (source slot) to San Francisco (destination slot). Table 1 shows mapping of X = $(X_1, \cdots, X_T)$ to the corresponding slot label Y = $(Y_1, \cdots, Y_T)$ in IOB format [9]. First, we have discussed Intent Classification and Slot Labelling; next, we discuss joint and parallel techniques of Intent and Slot Labelling. Lastly, we discuss about fusion in Visual Question Answering (VQA).

In a CA, there are multiple intents supported in a single domain. And given an intent, a particular set of slots are more likely than others. For example, once the system predicts that the intention is 'to fly', then the slots 'toLocation' and 'fromLocation' are more probable compared to others. But, sequential processing of Intent followed by Slot Prediction, is prone to additive error. Joint learning of both Intent and Slot Identification by fusion can reduce this additive error. Taking inspiration from Multimodal Compact Pooling fusion techniques [10] used in VQA, we have analyzed how fusion techniques can be applied to joint learning of Intent classification and Slot Labelling. For fusion, we have used Dense Addition and MLB fusion. We also posit that using Bidirectional GRUs instead of a CNN and Bi-LSTM will be able to better capture context within sentences and hence should improve the overall system performance.

TABLE 1. Sample Utterance from ATIS dataset representing semantic slots in IOB format

| Domain | Airline Travel |
|---|---|
| Intent | atis_flight |
| Sentence | Slot in IOB Format |
| Show | O |
| flights | O |
| from | O |
| Washington | B-fromloc.city_name |
| to | O |
| San | B-toloc.city_name |
| Francisco | I-toloc.city_name |
| between | O |
| 6 | B-depart_time.start_time |
| pm | I-depart_time.start_time |
| and | O |
| 8 | B-depart_time.end_time |
| pm | I-depart_time.end_time |
| on | O |
| friday | B-depart_date.day_name |

### A. Intent Classification

After the speech signal is converted into text by a Automatic Speech Recognition (ASR) system [11], identifying the intention of the user is called Intent Classification. Reference [12] addressed the problem of Intent Classification in a social media set up. They used a Hybrid Feature Representation method to handle data ambiguity. Reference [13] used three architectures of recurrent networks to perform multi-task learning for intent classification. Reference [14] proposed an ensemble of networks for intent classification, where outputs from GRU, Long Short Term Memory Network (LSTM) and CNN were collated using a Multi Layered Perceptron.

### B. Slot Labelling

Extracting the semantic values from an NL input is called Slot Labelling. Reference [15] used Recurrent Support Vector Machine (RSVM) for tagging slots. It was a two-step mechanism where the recurrent part first extracts input features. The SVM then evaluates a sequence-by-sequence objective function. The training phase is relatively fast, as weights of non-support vector samples were not updated.

LSTM Networks have often been utilized for sequence-based tasks like slot tagging. Reference [16] proposed an encoder-labeler LSTM. The first step was encoding of the input to a fixed size vector. This encoding was then used as an input for Slot Labeling, thereby capturing the global information from the input. Reference [17] proposed the use of a Deep Reinforcement Learning (DRL) based multimodal Coaching Model (DCM), where authors used a new reward mechanism for the RL based Slot Tagging. Even without any feedback from the user, this model was capable of understanding whatever has been wrongly labelled.

### C. Joint Intent and Slot Prediction

Since Intent Classification and Slot Tagging are highly correlated language understanding tasks, [13] proposed an attention based RNN method for this joint task, which utilized explicit alignment information in the encoder-decoder network. Reference [3] focused on better learning the relationship between the slot-intent entities, by using a slot-gated attention model, which predicts slots, based on the result of Intent.
Reference [18] did a comprehensive analysis of self-attention models, RNN and CNN to deal with the model obfuscation that arises from the joint Intent Classification (IC) and Slot Labelling (SL). They proposed a convolutional joint IC+SL technique for language understanding.

### D. Bi-Model or Parallel Intent and Slot Prediction

Most of the work on combined Intent and Slot prediction does the job sequentially or by using two separate models for individual tasks. However, [19] exploited the hierarchical relationship amongst intents and slots using Capsule Neural Networks with Dynamic Routing to learn the word features.

Similarly, [5] proposed a Bi-model Recurrent Semantic Parsing Network, which considers the impact of one task on the other. In this work, fusion technique is used to share knowledge between two models. Reference [20] proposed a technique with cross fusion of Slot labelling loss and Intent Classification loss while using a CNN for Intent Classification and BiLSTM-CRF for Slot Labelling

### E. Fusion in Visual Question Answering

In a Visual Question Answering (VQA) pipeline, the semantic interplay between the question and the image calls for feature fusion. Therefore, a fusion block takes the features from the question as well as the image and generates a multimodal feature output. Due to a similar relationship between the Intent Classification and Slot Labelling task, we used Multimodal Low Rank Bilinear fusion [21] for learning fused embedding for the NL input.

## II. PROPOSED WORK

We propose two parallel models to predict intent and slots while fusing their learnt information. A well-known technique of fusion is concatenation, which fails to capture complex contextual information. This encourages us to study and employ MLB and Dense Addition. In one experiment we applied CNN for intent and BiLSTM-CRF for slot prediction similar to [20], but with our proposed fusion techniques. In another experiment we replaced the architecture of [20] with Bidirectional GRU for both intent and slot prediction.

### A. Model-1 : CNN and Bi-LSTM for Intent and Slot Prediction using MLB and Dense Addition

For both Intent and Slot, we used same input sentence as shown in Figure 1. We converted the input sentence to a sequence of word vectors using Glove Embedding [22]. Glove Embedding provides 300 dimensional vectors (L1) for each seen word. We initialized a vector of 300 dimension to unseen words. Then we padded the input sentence to a max sequence length (L2), based on the type of data used as shown in Table 2. Hence, the resulting input dimensionality becomes L2 x L1. For Intent classification, a 4-layer parallel CNN architecture is used. We intended to capture unigram, bi-gram and higher n-gram features. For this, we used four different filter sizes 1 x L1, 2 x L1, 3 x L1 and 5 x L1 each representing the different number of words to be convolved in one go. 128 such filters were used at each layer. The extracted convolution features were then concatenated and passed to the dense layer.

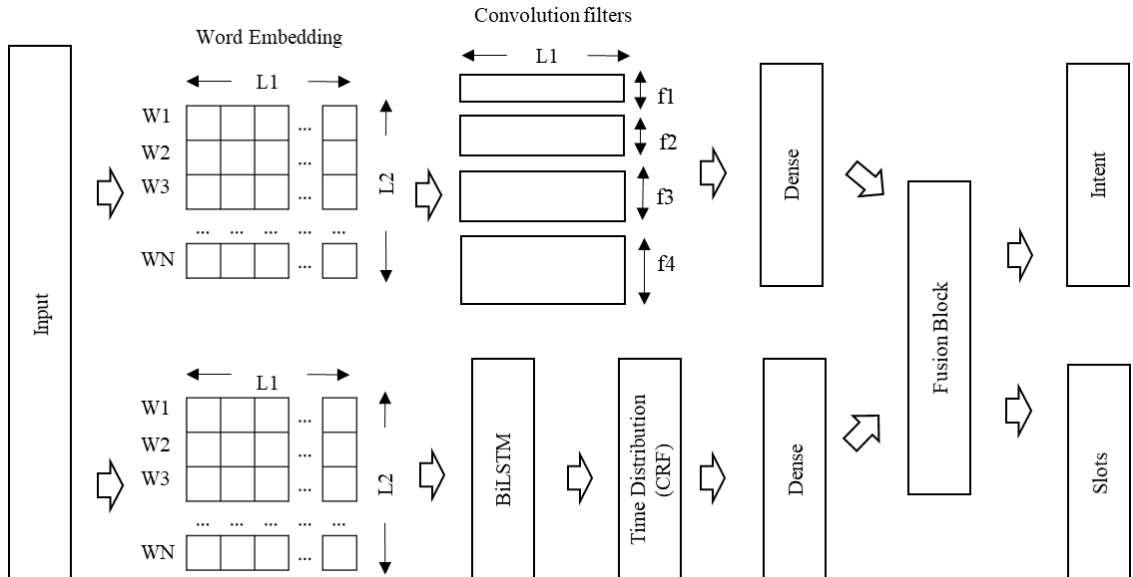

Figure 1: CNN and Bi-LSTM for Intent and Slot prediction using MLB and Dense Addition.

For slot prediction, the same embedding matrix of dimension L2 x L1 was passed to the Bi-LSTM model followed by single CRF layer. The output of CRF was passed on to the dense layer.

Following the dense layer, Intent output is fused with slot output using two different techniques: Dense Addition(Model-1a) and MLB Fusion(Model-1b). Further details of MLB and Dense Addition are shared in sub-sections *C* and *D*.

### B. Model-2 : Bidirectional GRU for Intent and Slot Prediction using MLB and Dense Addition

In Model-1, we are flattening the features and then passing them to subsequent layers. This leads to loss of word level semantic information. In this model, we use Bidirectional GRU instead of CNN to get better relationship of word. We also replaced Bi-LSTM with Bidirectional GRU because GRUs do not use a memory unit and are trained faster. They also perform at least as well as LSTM [22] and are easy to modify when new input features need to be added to the network.

We used same input sentence as used in Model-1 and converted the input sentence to sequence of word embedding using Glove Embedding [23]. In Intent Classification, we fed the above word embedding sequence to a Bi-Directional GRU that captured the contextual information from both directions. It used this information to predict the output for a particular sentence. We used 128 dimensions as output features for each GRU. We concatenated the output from each of them in the feature axis. The resulting dimension was L2 x 256. We used 0.5 as dropout followed by a dense layer of 128 dimension.

For Slot Classification, we fed the word embedding sequence to the Bi-Directional GRU, which provided contextual information from both the directions and predicted the output of the word. We took sequence output from each GRU and concatenated it, as done in intent classification to get final feature representation. The final dimensional information was L2 x 256. Then we passed these features through a Dense layer to get 128 feature learnings.

Similar to Model-1, intent output was fused with slot output. Dense Addition(Model-2a) and MLB Fusion(Model-2b) were used for performing this fusion. Finally, for Intent Prediction, we flattened the fused learnings and passed it through dense layer with Softmax activation. Similarly, for Slot prediction, the fused learnings were passed through Dense layer with Softmax activation to obtain final Slot Predictions.

### C. MLB Fusion

A fusion block as shown in Figure 3, is used to combine the learnings of two separate models through multiplication with the weight matrix.

Given, input vectors x and y, where x represents the output for intent and y represents slot output, we get the fused learning $f_i$ as:

$$f_i = x^T W_i y + b_i \qquad (1)$$

Where $W_i$ is the weight matrix of order (m*n), and $b_i$ is the bias term for $f_i$. This means that there are a total of l*(m*n+1) features with 'l' being the number of output features. This is computationally very expensive. To lower this large number of parameters to be learnt, we used MLB.

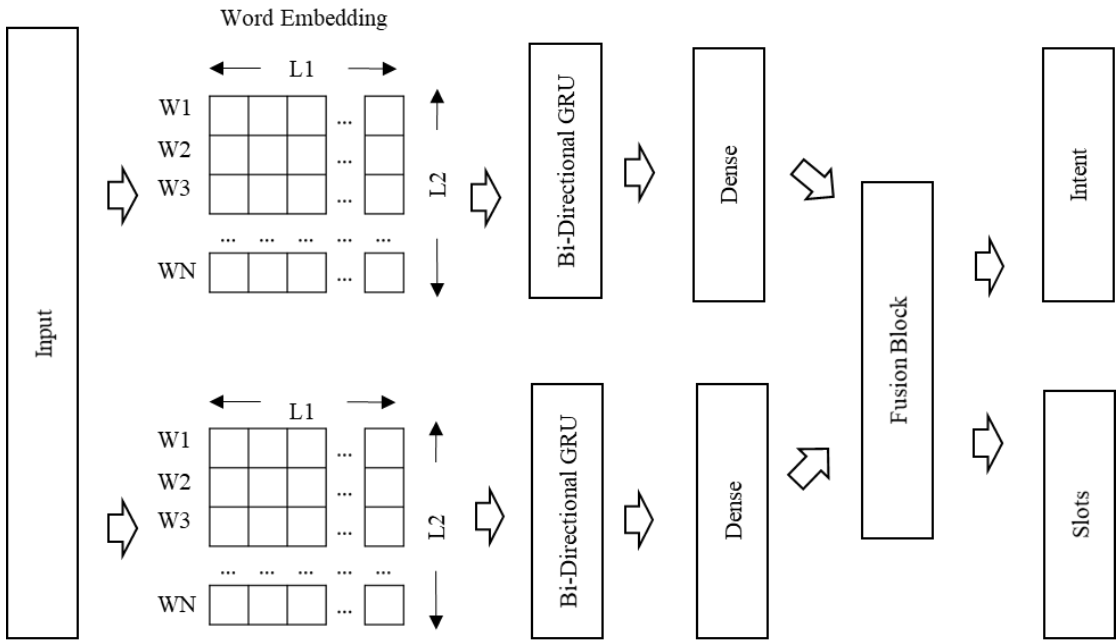

Figure 2: Bidirectional GRU for Intent and Slot prediction using MLB and Dense Addition.

MLB technique addresses the problem of high-dimensional data by decomposing the weight matrix [21] $W_i$ of order (m*n) into two smaller matrices $U_i$ and $V_i$ of order (m*k) and (n*k) respectively, with k less than min(m, n), such that:

$$W_i = U_i V_i^T \quad (2)$$

The rank of a matrix M of order (a*b) is limited by min(a, b). So, by performing this decomposition of W into U and V, we limit the "rank" of the $W_i$ to be at most 'k', which is less than minimum of m and n. This reduces the computational complexity of the system. Using equation (2) in (1) further leads to:

$$f_i = x^T W_i y + b_i$$
$$= x^T U_i V_i^T y + b_i$$
$$= 1^T (U_i^T x \circ V_i^T y) + b_i \quad (3)$$

Where ∘ denotes the Hadamard Product and $1^T$ is a column vector of ones.

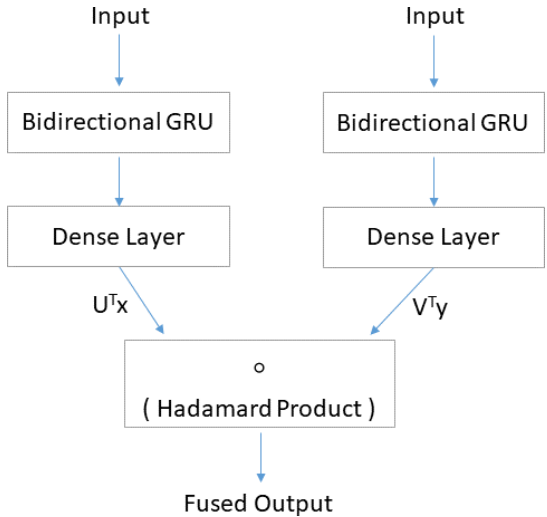

Figure 3: MLB Fusion

### D. Dense Addition

Reference [20] used three different techniques for fusion - concatenation, addition and average. In their experiment, they used a CNN for the intent task, which gave a two-dimensional output, and a Bi-LSTM for slots, which gave three-dimensional output. In order to fuse both they proposed reshaping the two-dimensional output by using broadcast to convert it into three dimensions.

In our experiment, since we are using Bidirectional GRU for both the tasks we simply added the learnings from intent and slots. The output of the addition of the two was used for final prediction of intent and slots.

### III. DATASETS AND EXPERIMENTAL DETAILS

The proposed architectures were evaluated on two benchmark datasets for Intent and Slot prediction, 'ATIS' and 'Snips'. Both these datasets have been taken from the GitHub Source mentioned with Table 2.

**ATIS Dataset:** The Airline Travel Information Systems dataset consists of user-spoken utterances for flight reservation. There are a total of 4,978 training utterances, 893 test utterances and 500 utterances as validation data. The total number of intent classes to be predicted is 21 and total number of unique slots is 120. The maximum length of each input sentence is fixed at 50.

**Snips Dataset:** This data has been collected using the Snips Personal Voice Assistant. In Snips data, each intent is uniformly distributed. The training set consists of 13,084 utterances, while the validation and test set have 700 utterances each. Total number of unique intent labels present are 7, while there are 72 unique slots. The maximum sentence length is 36.

TABLE 2. ATIS and Snips datasets used in experiment

| Datasets | ATIS[1] | Snips[2] |
|---|---|---|
| Train Data | 4,978 | 13,084 |
| Test Data | 893 | 700 |
| Validation Data | 500 | 700 |
| Vocabulary Size | 722 | 11,241 |
| Slots | 120 | 72 |
| Intents | 21 | 7 |
| Maximum Sentence Length | 50 | 36 |

The architecture as depicted in Figure 1 and Figure 2 were built using Keras framework. The training was set up to 100 epochs although the models converged earlier. The loss function used here was "categorical_cross_entropy" with "adam" optimizer and a batch size of 64.

### IV. EVALUATION AND DISCUSSION

We have evaluated the proposed architectures on two open source datasets using Glove embedding. We first discuss the impact of different fusion techniques (Model-1a and Model-1b) on CNN + Bi-LSTM architecture. Next, we study the improvement by using Bidirectional GRU(Model-2a and Model-2b) instead of CNN + Bi-LSTM. Lastly, we do a holistic comparison of our best architecture with the current-state-of-the-art.

### A. Impact of MLB and Dense Addition

In Model-1a and Model-1b, we experimented with Dense Addition and MLB fusion respectively. Table 3 shows the impact of using MLB and Dense addition instead of simple cross fusion techniques as used by [20]. With these two fusion

---
[1] https://github.com/yvchen/JointSLU/tree/master/data
[2] https://github.com/MiuLab/SlotGated-SLU

techniques, we are able to match and even surpass state-of-the-art results on intent and slots.

TABLE 3. Intent Accuracy & Slot F1-score of Model-1 on two benchmark datasets with MLB and Dense Addition

| Technique | ATIS | | Snips | |
|---|---|---|---|---|
| | Intent | Slot | Intent | Slot |
| Bhasin et al. [20] | 97.42 | 99.54 | 98.14 | 98.44 |
| CNN/Bi-LSTM with Dense Addition (Model-1a) | 97.53 | 99.47 | 94.14 | 98.44 |
| CNN/Bi-LSTM with MLB Fusion (Model-1b) | **97.54** | **99.54** | **98.14** | **98.49** |

### B. Impact of using Bidirectional GRU with MLB Fusion and Dense Addition

In Model-2, we used Bidirectional GRUs with Dense Addition (Model-2a) and MLB Fusion (Model-2b). A GRU has just two gates, reset gate and update gate while an LSTM has input, output and forget gates. Replacing Bi-LSTM with Bidirectional GRU helped to train the model faster.

Model-2b further improved our results from Model-1b by 0.22% and 0.06% on ATIS intent and slot respectively, as well as 0.28% and 0.25% on Snips. Table 4 shows the impact of the same.

TABLE 4. Intent Accuracy and Slot F1-score obtained with Model-2 compared with Model-1

| Technique | ATIS | | Snips | |
|---|---|---|---|---|
| | Intent | Slot | Intent | Slot |
| CNN/Bi-LSTM with Dense Addition (Model-1a) | 97.53 | 99.47 | 94.14 | 98.44 |
| CNN/Bi-LSTM with MLB Fusion (Model-1b) | 97.54 | 99.54 | 98.14 | 98.49 |
| Bidirectional GRU with Dense Addition (Model-2a) | 97.65 | 99.56 | 98.14 | 98.44 |
| Bidirectional GRU with MLB Fusion (Model-2b) | **97.76** | **99.60** | **98.42** | **98.74** |

### C. Overall comparison with other state-of-the-art techniques

In this section, we have compared our overall best model, Model-2b, which has Bidirectional GRUs with MLB fusion with the latest works [3-4] and current state-of-the-art [20]. Table 5 shows comparison of the same.

TABLE 5. Comparison of Intent Accuracy and Slot F1-score with other state-of-the-art models

| Technique | ATIS | | Snips | |
|---|---|---|---|---|
| | Intent | Slot | Intent | Slot |
| Goo et al. [3] (Full Attention) | 93.6 | 94.8 | 97.0 | 88.8 |
| Goo et al. [3] (Intent Attention) | 94.1 | 95.2 | 96.8 | 88.3 |
| Wang et al. [4] | 97.17 | 97.76 | - | - |
| Bhasin et al. [20] | 97.42 | 99.54 | 98.14 | 98.44 |
| Model-2b (Best) | **97.76** | **99.60** | **98.42** | **98.74** |

With Bidirectional GRU and MLB fusion technique, we were able to surpass state-of-the-art results by 0.34% in ATIS intent classification accuracy and 0.28% in case of Snips. Similarly, for slot prediction, we obtained an improvement of 0.30% F1-score in case of snips and a marginal improvement of 0.06% in ATIS.

The above results prove our hypothesis that using Bidirectional GRUs with fusion instead of CNN + Bi-LSTM will be better in capturing context and lead to improvement in Intent Accuracy and Slot F1-score. We also observe, that MLB fusion outperforms Dense addition in most cases.

## V. CONCLUSION

The proposed Bidirectional GRU model with Fusion obtained state-of-the-art results. The accuracies for both Intent Classification and Slot Labelling were improved. By utilizing the MLB technique, we addressed the problem of computation of large number of weights in a matrix. Since the state-of-the-art is already in the high 90's, there is a limited scope of further improvement in this task, but we posit that this technique can be applied to other multi-task learning problems in the domain of Natural Language like Domain Classification or Named Entity Recognition. We also encourage the fellow research community to pursue the same.